\documentclass[10pt,twocolumn,letterpaper]{article}
\usepackage[dvipsnames]{xcolor}
\usepackage[pagenumbers]{cvpr} 
\definecolor{cvprblue}{rgb}{0.21,0.49,0.74}
\usepackage[pagebackref,breaklinks,colorlinks,citecolor=cvprblue]{hyperref}
\usepackage{array}
\usepackage{setspace}

\def\paperID{*****} 
\def\confName{CVPR}
\def\confYear{2024}

\title{Exploring the Spectrum of Visio-Linguistic Compositionality and Recognition
\vspace{-2mm}}

\newcommand{\mailto}[2]{\href{mailto:#1}{\color{black}#2}}
\newcommand\coauthormark{\footnotemark[\arabic{footnote}]}
\author{Youngtaek Oh\textsuperscript{1}\thanks{This work was done during a research internship at LG AI Research.} \qquad Pyunghwan Ahn\textsuperscript{2} \qquad Jinhyung Kim\textsuperscript{2} \qquad Gwangmo Song\textsuperscript{2} \vspace{1mm}\\
Soonyoung Lee\textsuperscript{2}\thanks{Corresponding authors} \qquad\quad In So Kweon\textsuperscript{1}\protect\coauthormark \qquad\quad Junmo Kim\textsuperscript{1}\protect\coauthormark\vspace{1mm}\\
  \textsuperscript{1}KAIST \qquad\qquad\textsuperscript{2}LG AI Research\\
  \small{
      \textsuperscript{1}\texttt{\{\mailto{youngtaek.oh@kaist.ac.kr}{youngtaek.oh}, \mailto{iskweon77@kaist.ac.kr}{iskweon77}, \mailto{junmo.kim@kaist.ac.kr}{junmo.kim}\}@kaist.ac.kr
      }
  }\\
  \small{
    \textsuperscript{2}\texttt{\{\mailto{p.ahn@lgresearch.ai}{p.ahn}, \mailto{jinhyung.kim@lgresearch.ai}{jinhyung.kim}, \mailto{gwangmo.song@lgresearch.ai}{gwangmo.song}, \mailto{soonyoung.lee@lgresearch.ai}{soonyoung.lee}\}@lgresearch.ai}
  }
}

\begin{document}
\maketitle
\begin{abstract}
Vision and language models (VLMs) such as CLIP have showcased remarkable zero-shot recognition abilities yet face challenges in visio-linguistic compositionality, particularly in linguistic comprehension and fine-grained image-text alignment. This paper explores the intricate relationship between compositionality and recognition -- two pivotal aspects of VLM capability. We conduct a comprehensive evaluation of existing VLMs, covering both pre-training approaches aimed at recognition and the fine-tuning methods designed to improve compositionality. Our evaluation employs 12 benchmarks for compositionality, along with 21 zero-shot classification and two retrieval benchmarks for recognition. In our analysis from 274 CLIP model checkpoints, we reveal patterns and trade-offs that emerge between compositional understanding and recognition accuracy. Ultimately, this necessitates strategic efforts towards developing models that improve both capabilities, as well as the meticulous formulation of benchmarks for compositionality. We open our evaluation framework at \url{https://github.com/ytaek-oh/vl_compo}.
\end{abstract}

\section{Introduction}
\label{sec:intro}
The advent of vision and language models (VLMs) like CLIP~\cite{radford2021learning} has significantly advanced artificial intelligence by merging visual and textual data, showcasing exceptional zero-shot recognition abilities for identifying previously unseen objects. Despite the strong recognition ability, recent studies~\cite{yuksekgonul2023when, ma2023crepe, thrush2022winoground} have uncovered their poor ability of compositional reasoning: identifying objects in an image while also understanding their complex relationships and contexts along with the accompanying text. As such, visio-linguistic compositionality emerges as another crucial dimension in the capabilities of VLMs alongside recognition. Despite both being as essential axes for VLMs, research has traditionally approached them in isolation, overlooking their interconnected impact on VLMs. 
The effects of standard pre-training methods aimed at boosting recognition~\cite{mu2022slip,li2022supervision} on compositionality remains less explored.
Conversely, models focused on fine-tuning for compositionality often lose the zero-shot recognition ability~\cite{zhang2023contrasting,doveh2023dense}. This bifurcation suggests the necessity for a more integrated evaluation framework for both types of VLMs, facilitating a deeper understanding of VLM capabilities.

\begin{figure}[t]
  \centering
   \includegraphics[width=0.90\columnwidth]{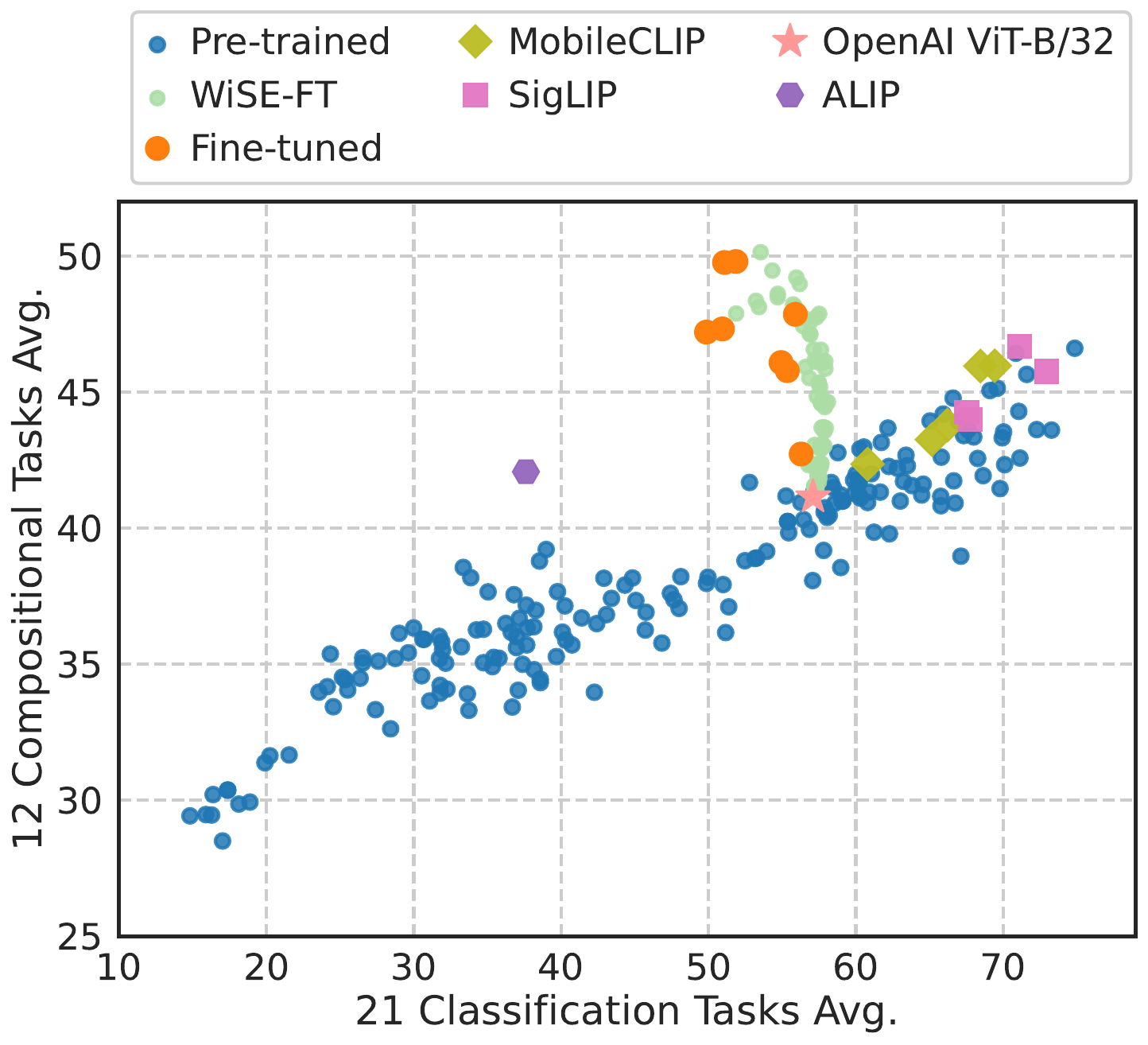}
   \vspace{-2mm}
   \caption{A comprehensive overview of the trend between compositionality and recognition. Pre-trained VLMs (in blue points) exhibit improved compositionality alongside enhanced zero-shot classification. Conversely, models fine-tuned for compositionality (in green and orange points) demonstrate trade-offs between these two capabilities. A detailed analysis is provided in~\cref{sec:experiments}.
   }
   \label{fig:teaser}
   \vspace{-3mm}
\end{figure}

Pursuing this direction, we embark on a comprehensive co-evaluation of existing VLMs, with a focus on both compositionality and zero-shot recognition tasks.  Our evaluation, as outlined in~\cref{fig:teaser}, spans a wide array of benchmarks, including 12 for compositionality, 21 for zero-shot classification, and additionally, 2 for zero-shot retrieval tasks. We leverage a broad spectrum of publicly available CLIP model checkpoints: from those pre-trained on large-scale datasets~\cite{ilharco_gabriel_2021_5143773,vasu2023mobileclip,mu2022slip,goel2022cyclip,desai2023meru,li2022supervision} to those fine-tuned specifically for compositionality~\cite{yuksekgonul2023when,doveh2023dense,zhang2023contrasting}, and even models that merge the advantages of zero-shot and fine-tuned methods using weight-space ensembling (WiSE-FT)~\cite{wortsman2022robust}. 
Our aim is to jointly evaluate and establish a comprehensive benchmark for VLMs, covering the two capabilities.

\begin{figure}[t]
  \centering
   \includegraphics[width=0.85\linewidth]{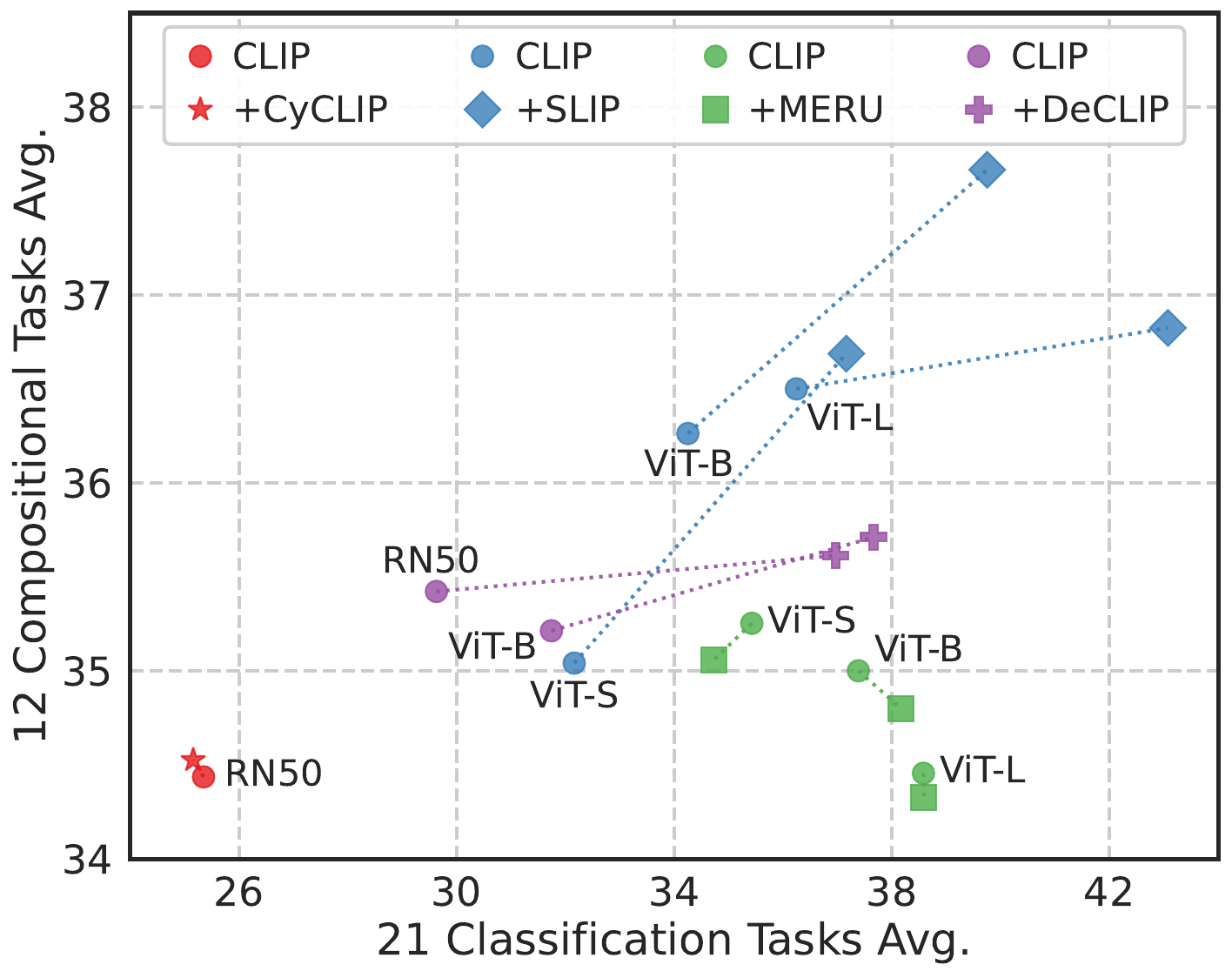}
   \vspace{-2mm}
   \caption{Nuanced trade-offs between compositionality and recognition in VLMs. While enhanced pre-training methods can lead to improvements in both areas (\eg, SLIP and DeCLIP), this is not always the case. \cref{sec:pretrained} provides a detailed explanation.}
   \label{fig:small_pretrained}
   \vspace{-3mm}
\end{figure}

As outlined in~\cref{sec:method}, we developed a toolkit to streamline the coherent evaluation of VLMs across these dimensions. Featuring a unified codebench, it integrates all benchmarks and models in our study, thereby facilitating an efficient evaluation. Importantly, our framework is built with scalability in mind, designed to accommodate additional benchmarks and models, ensuring its relevance and utility in ongoing research efforts across the community.

From~\cref{sec:experiments}, utilizing a diverse collection of models, our evaluation is organized into several distinct sections. Initially, in~\cref{sec:pretrained}, we explore the trends between compositionality and recognition among pre-trained models, taking into account variations in data and model scales, alongside models featuring specialized pre-training objectives. Subsequently, in~\cref{sec:finetuned}, we present a detailed analysis of models fine-tuned for compositionality, evaluating their effectiveness across a range of recognition tasks, including zero-shot classification and retrieval.

\begin{figure}[t]
  \centering
   \includegraphics[width=0.95\linewidth]{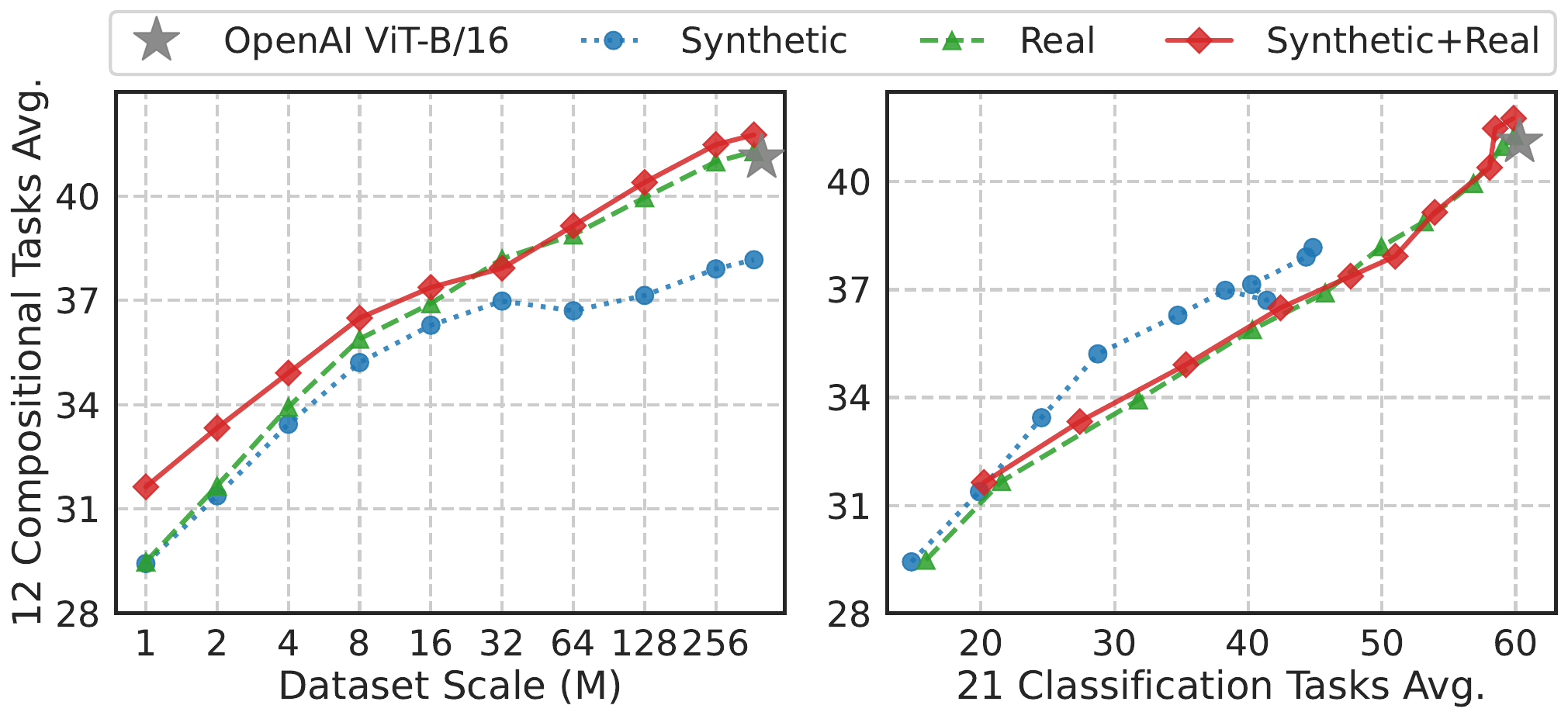}
   \vspace{-2mm}
   \caption{Data scaling property in compositionality tasks. (Left) Compositionality improves along with the scale of pre-training data. (Right) Pre-training with data that include real images tends to yield better efficiency in compositionality against recognition.}
   \label{fig:scaling}
   \vspace{-2mm}
\end{figure}

In summary, our contributions are as follows:
\begin{itemize}

\item \textbf{Comprehensive Evaluation Suite for VLMs}:
We establish a benchmark that evaluates both compositionality and zero-shot recognition, providing extensive results in a wide range of VLMs to illuminate these two capabilities.

\item \textbf{Understanding VLMs capabilities}:
We shed light on the nuanced dynamics and trade-offs between compositionality and zero-shot recognition in the realm of VLMs, laying a groundwork for subsequent progress in the field.

\end{itemize}

\begin{table}[t]
\centering
\resizebox{.90\linewidth}{!}{%
    \begin{tabular}{p{2.5cm} p{7.5cm}}
    \toprule
        Task  &  Benchmarks  \\
    \midrule
        Compositionality & ARO, CREPE, SugarCrepe, VALSE, VL Checklist, WhatsUp, ImageCoDe, SVO Probes, Winoground, ColorSwap, EqBen, MMVP-VLM \\
    \midrule
        Retrieval & Flickr30k, COCO \\
    \midrule
        Classification & ImageNet, ELEVATOR \\
    \bottomrule
\end{tabular}
}
\vspace{-1mm}
\caption{A list of benchmarks in our evaluation toolkit.
}
\label{tab:benchmarks}
\vspace{-4mm}
\end{table}

\begin{figure*}[t]
  \centering
   \includegraphics[width=0.90\linewidth]{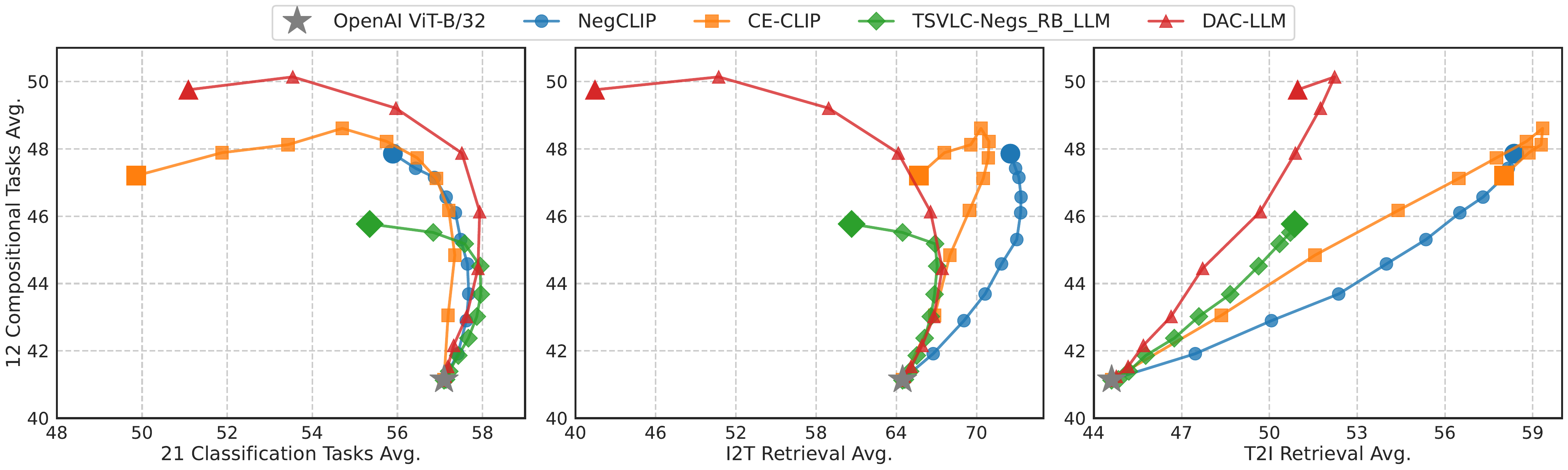}
   \vspace{-2mm}
   \caption{Exploring fine-tuning effects on compositionality and recognition through the lens of weight-space ensembling (WiSE-FT~\cite{wortsman2022robust}). (Left) Increased compositionality comes at the cost of zero-shot classification accuracy. For retrieval tasks, (Center) fine-tuned models with COCO (\eg, NegCLIP and CE-CLIP) enhances I2T recall in the initial stages, whereas fine-tuning with datasets less akin to COCO, such as CC3M (\eg, TSVLC and DAC), can result in noticeable drops. (Right) Consistent across all models, fine-tuning benefits T2I recall.}
   \label{fig:wiseft}
   \vspace{-3mm}
\end{figure*}

\section{Evaluation Toolkit}
\label{sec:method}
We introduce a toolkit for evaluating compositionality and zero-shot recognition of VLMs. As presented in~\cref{tab:benchmarks}, our toolkit incorporates 12 compositionality benchmarks and zero-shot recognition tasks including classification and retrieval. This toolkit is compatible with \textit{open\_clip}~\cite{ilharco_gabriel_2021_5143773} models, and also incorporates diverse publicly available models, offering a broad support. While \textit{CLIP\_benchmark} project~\cite{cherti2024clip} focuses on zero-shot recognition, linear probing, and multilingual support, we place a greater emphasis on compositionality tasks. We also emphasize that our toolkit is designed for scalability, allowing it to easily incorporate additional benchmarks and models in the future.

\noindent \textbf{Compositionality Benchmarks.} In this task, given either a query image or text, VLMs are tasked to select the correct match of text or image from a candidate set including subtly manipulated incorrect options. 
To illustrate, examples of textual variations include manipulations in spatial relations~\cite{yuksekgonul2023when}, attributes~\cite{yuksekgonul2023when,hsieh2024sugarcrepe}, and negations~\cite{ma2023crepe}, showcasing the breadth in probing compositional understanding. 

We categorize compositionality benchmarks into three: (1) Image-to-Text (I2T), where the task involves selecting the correct text match for a given image from among negative options such as ARO~\cite{yuksekgonul2023when} and SugarCrepe~\cite{hsieh2024sugarcrepe}. (2) Text-to-Image (T2I), the reversed scenario of I2T, with the goal being to identify the correct image from a set of negative images based on a given text query, like SVO Probes~\cite{hendricks-nematzadeh-2021-probing}. (3) A \textit{Group} setting, merging I2T and T2I, provides two sets of matched image-text pairs for a single test, and the model is required to correctly associate each image with its text and the vice versa. This approach is illustrated by benchmarks such as Winoground~\cite{thrush2022winoground}.

For I2T and T2I tasks, we utilize top-1 accuracy for evaluation. In the Group setting, we employ group accuracy to  identify all correct image-text matches and their inverse for a single test.
For each compositionality benchmark with multiple tasks and their subtasks, we recursively group and average scores of subtasks linked by a common ancestor, progressively consolidating them into a single average score. We use this as the evaluation metric for a benchmark.

\noindent \textbf{Zero-shot Recognition Benchmarks.}
For zero-shot classification, we utilize a combination of the ImageNet~\cite{deng2009imagenet} validation split and the ELEVATOR toolkit~\cite{NEURIPS2022_elevator} covering 21 datasets in total, following previous literature~\cite{doveh2023teaching,doveh2023dense}. For zero-shot retrieval, we use the COCO karpathy~\cite{karpathy2015deep} and the Flickr30k datasets. We report top-1 accuracy for classification and Recall@1 for both image and text in retrieval tasks.

\noindent \textbf{Models.}
We focus on contrastive VLMs with variations in data and model scales, objectives, embeddings, and architecture. 
We have meticulously collected model checkpoints for joint evaluations on recognition and compositionality. This collection includes 194 pre-trained models, and 8 models fine-tuned for compositionality based the OpenAI CLIP ViT-B/32 model.
Additionally, 72 models are produced by weight-space ensembling (WiSE-FT)~\cite{wortsman2022robust}, which is known to make better trade-offs between OOD and ID accuracy. We adjust the blending weight $\alpha$ from 0.0 (pre-trained) to 1.0 (fully fine-tuned) in 0.1 increments to create 9 intermediate models for each fine-tuned variant. We apply this across all fine-tuned models to examine the trade-offs between compositional and recognition tasks.

\section{Experiments}
\label{sec:experiments}
We evaluate diverse pre-trained and fine-tuned CLIP models jointly on compositionality and recognition tasks coherently, presenting several key findings. As note, \cref{sec:sup_eval_details} contains a comprehensive list of benchmarks and models utilized in our experiments, and \cref{sec:sup_detailed_res} contains an expanded analysis for a compositionality benchmark.

\subsection{Analysis on Pre-trained Models}
\label{sec:pretrained}
\noindent \textbf{Better recognition leads to improved compositionality.} 
As presented in~\cref{fig:teaser}, there is a positive correlation between compositionality and zero-shot classification tasks. This indicates that compositional reasoning skills improves along with the recognition performance. 
Within this group, SigLIP~\cite{zhai2023sigmoid}, a strong recognition model, exhibits superior compositional abilities. 
Moreover, despite their compact sizes, ALIP~\cite{yang2023alip} employing synthetic captions with an adaptive loss, and MobileCLIP~\cite{vasu2023mobileclip} benefiting from dataset reinforce, display remarkable trade-offs.

\noindent \textbf{Improved pre-training objectives do not guarantee enhanced compositionality.}
We examine a series of CLIP models with advanced pre-training methodologies, including in objectives~\cite{goel2022cyclip,mu2022slip,li2022supervision} and representation space~\cite{desai2023meru}. 
It is noteworthy that all these models were pre-trained on relatively small-scale datasets, such as CC3M and YFCC15M. 
From~\cref{fig:small_pretrained}, although SLIP~\cite{mu2022slip} and DeCLIP~\cite{li2022supervision} improved both compositionality and recognition, this trend was not observed across other models. This highlights the need for further exploration of pre-training methodologies for CLIP that could also help compositionality.

\noindent \textbf{Compositionality arises with data scaling.}
We examine data scaling effects on compositionality, utilizing CLIP models pre-trained on subsets of LAION-400M ranging from 1M to 371M, acquired from~\cite{fan2023scaling}. 
As shown in~\cref{fig:scaling}, there is a clear correlation between the scale of data and compositionality. 
Our analysis indicates that models pre-trained solely on synthetic images (highlighted by blue lines) exhibit lower efficiency, achieving less recognition accuracy than models trained with real samples (marked by green and red lines) to attain comparable levels of compositionality.
It also shows that pre-training with either purely real images or a combination of real and synthetic images results in superior recognition and compositionality compared to OpenAI CLIP ViT-B/16 trained on 400M samples.
 
\subsection{Analysis on Fine-tuned Models}
\label{sec:finetuned}
We explore fine-tuned models designed to enhance compositionality, each applied to the OpenAI CLIP ViT-B/32 and utilizing objectives that include hard negatives~\cite{yuksekgonul2023when,zhang2023contrasting,doveh2023teaching,doveh2023dense}. Additionally, we apply WiSE-FT~\cite{wortsman2022robust} to examine the fine-tuning trajectories starting from the pre-trained model with respect to the performances, as presented in~\cref{fig:wiseft}. 

\noindent \textbf{Clear trade-offs with fine-tuning.} 
For zero-shot classification tasks, as shown in~\cref{fig:wiseft} (left), fully fine-tuned models gain compositionality at the expense of recognition accuracy, suggesting loss of the pre-trained model's inherent knowledge during fine-tuning. Although both TSVLC and DAC use LoRA~\cite{hu2022lora} to preserve original weights during fine-tuning, it still faces this degradation. Meanwhile, at intermediate stages, there are periods where both compositionality and recognition improve, consistent with the observations made in WiSE-FT~\cite{wortsman2022robust}. A training scheme that can retain recognition ability represents a desirable direction.

\noindent \textbf{Mixed trade-offs in retrieval tasks.}
We observe that the nature of the trade-offs depends on the fine-tuning dataset. As shown in~\cref{fig:wiseft} (center), 
NegCLIP and CE-CLIP, fine-tuned on COCO, showed noticeable gain in I2T recall (\eg, averaged across COCO and Flickr30k) during fine-tuning. Conversely, TSVLC and DAC, fine-tuned on CC3M, which is less akin to COCO, experienced minimal improvements or even severe declines in I2T recall as fine-tuning progressed.
We speculate that this disparity stems from the shared data characteristics between the training and evaluation datasets.
We note that to ensure an unbiased evaluation of VLMs, it is essential to avoid any direct knowledge transfer from the training data to the evaluation phase, as also shared in~\cite{singh-etal-2023-coarse}. 
As such, we believe that an evaluation task independent from training data would be useful. For T2I retrieval tasks from~\cref{fig:wiseft} (right), fine-tuning for compositionality consistently improves T2I recall across all models.

\section{Conclusion}
We investigated the intricate yet underexplored relationship between visio-linguistic compositionality and recognition tasks within vision and language models (VLMs). Through our extensive evaluation, which includes both pre-trained models for recognition and models fine-tuned for compositionality, we highlight the necessity of strategies that concurrently enhance both capabilities. The benchmarking results alongside our evaluation framework provide a comprehensive perspective that lays the groundwork for future advancements in VLMs, with the goal of enhancing their ability to understand and interact with the visual and linguistic aspects of the world.

{
    \small
    \bibliographystyle{ieeenat_fullname}
    \bibliography{main}

\begin{thebibliography}{42}
\providecommand{\natexlab}[1]{#1}
\providecommand{\url}[1]{\texttt{#1}}
\expandafter\ifx\csname urlstyle\endcsname\relax
  \providecommand{\doi}[1]{doi: #1}\else
  \providecommand{\doi}{doi: \begingroup \urlstyle{rm}\Url}\fi

\bibitem[Burapacheep et~al.(2024)Burapacheep, Gaur, Bhatia, and Thrush]{burapacheep2024colorswap}
Jirayu Burapacheep, Ishan Gaur, Agam Bhatia, and Tristan Thrush.
\newblock Colorswap: A color and word order dataset for multimodal evaluation.
\newblock \emph{arXiv preprint arXiv:2402.04492}, 2024.

\bibitem[Castro et~al.(2024)Castro, Ziai, Saluja, Yuan, and Mihalcea]{castro2024clove}
Santiago Castro, Amir Ziai, Avneesh Saluja, Zhuoning Yuan, and Rada Mihalcea.
\newblock Clove: Encoding compositional language in contrastive vision-language models.
\newblock \emph{arXiv preprint arXiv:2402.15021}, 2024.

\bibitem[Cherti et~al.(2022)]{cherti2024clip}
Mehdi Cherti et~al.
\newblock Clip\_benchmark, 2022.
\newblock \url{https://github.com/LAION-AI/CLIP_benchmark}.

\bibitem[Cui et~al.(2022)Cui, Zhao, Liang, Li, and Shao]{cui2022democratizing_declip}
Yufeng Cui, Lichen Zhao, Feng Liang, Yangguang Li, and Jing Shao.
\newblock Democratizing contrastive language-image pre-training: A clip benchmark of data, model, and supervision, 2022.

\bibitem[Deng et~al.(2009)Deng, Dong, Socher, Li, Li, and Fei-Fei]{deng2009imagenet}
Jia Deng, Wei Dong, Richard Socher, Li-Jia Li, Kai Li, and Li Fei-Fei.
\newblock Imagenet: A large-scale hierarchical image database.
\newblock In \emph{2009 IEEE conference on computer vision and pattern recognition}, pages 248--255. Ieee, 2009.

\bibitem[Desai et~al.(2023)Desai, Nickel, Rajpurohit, Johnson, and Vedantam]{desai2023meru}
Karan Desai, Maximilian Nickel, Tanmay Rajpurohit, Justin Johnson, and Shanmukha~Ramakrishna Vedantam.
\newblock Hyperbolic image-text representations.
\newblock In \emph{Proceedings of the 40th International Conference on Machine Learning}, pages 7694--7731. PMLR, 2023.

\bibitem[Doveh et~al.(2023{\natexlab{a}})Doveh, Arbelle, Harary, Herzig, Kim, Cascante-Bonilla, Alfassy, Panda, Giryes, Feris, et~al.]{doveh2023dense}
Sivan Doveh, Assaf Arbelle, Sivan Harary, Roei Herzig, Donghyun Kim, Paola Cascante-Bonilla, Amit Alfassy, Rameswar Panda, Raja Giryes, Rogerio Feris, et~al.
\newblock Dense and aligned captions (dac) promote compositional reasoning in vl models.
\newblock \emph{Advances in Neural Information Processing Systems}, 36, 2023{\natexlab{a}}.

\bibitem[Doveh et~al.(2023{\natexlab{b}})Doveh, Arbelle, Harary, Schwartz, Herzig, Giryes, Feris, Panda, Ullman, and Karlinsky]{doveh2023teaching}
Sivan Doveh, Assaf Arbelle, Sivan Harary, Eli Schwartz, Roei Herzig, Raja Giryes, Rogerio Feris, Rameswar Panda, Shimon Ullman, and Leonid Karlinsky.
\newblock Teaching structured vision \& language concepts to vision \& language models.
\newblock In \emph{Proceedings of the IEEE/CVF Conference on Computer Vision and Pattern Recognition}, pages 2657--2668, 2023{\natexlab{b}}.

\bibitem[Fan et~al.(2023)Fan, Chen, Krishnan, Katabi, Isola, and Tian]{fan2023scaling}
Lijie Fan, Kaifeng Chen, Dilip Krishnan, Dina Katabi, Phillip Isola, and Yonglong Tian.
\newblock Scaling laws of synthetic images for model training... for now.
\newblock \emph{arXiv preprint arXiv:2312.04567}, 2023.

\bibitem[Fan et~al.(2024)Fan, Krishnan, Isola, Katabi, and Tian]{fan2024improving_laclip}
Lijie Fan, Dilip Krishnan, Phillip Isola, Dina Katabi, and Yonglong Tian.
\newblock Improving clip training with language rewrites.
\newblock \emph{Advances in Neural Information Processing Systems}, 36, 2024.

\bibitem[Goel et~al.(2022)Goel, Bansal, Bhatia, Rossi, Vinay, and Grover]{goel2022cyclip}
Shashank Goel, Hritik Bansal, Sumit Bhatia, Ryan Rossi, Vishwa Vinay, and Aditya Grover.
\newblock Cyclip: Cyclic contrastive language-image pretraining.
\newblock \emph{Advances in Neural Information Processing Systems}, 35:\penalty0 6704--6719, 2022.

\bibitem[Hammoud et~al.(2024)Hammoud, Itani, Pizzati, Torr, Bibi, and Ghanem]{hammoud2024synthclip}
Hasan Abed Al~Kader Hammoud, Hani Itani, Fabio Pizzati, Philip Torr, Adel Bibi, and Bernard Ghanem.
\newblock Synthclip: Are we ready for a fully synthetic clip training?
\newblock \emph{arXiv preprint arXiv:2402.01832}, 2024.

\bibitem[Hendricks and Nematzadeh(2021)]{hendricks-nematzadeh-2021-probing}
Lisa~Anne Hendricks and Aida Nematzadeh.
\newblock Probing image-language transformers for verb understanding.
\newblock In \emph{Findings of the Association for Computational Linguistics: ACL-IJCNLP 2021}, pages 3635--3644, Online, 2021. Association for Computational Linguistics.

\bibitem[Hsieh et~al.(2024)Hsieh, Zhang, Ma, Kembhavi, and Krishna]{hsieh2024sugarcrepe}
Cheng-Yu Hsieh, Jieyu Zhang, Zixian Ma, Aniruddha Kembhavi, and Ranjay Krishna.
\newblock Sugarcrepe: Fixing hackable benchmarks for vision-language compositionality.
\newblock \emph{Advances in Neural Information Processing Systems}, 36, 2024.

\bibitem[Hu et~al.(2022)Hu, yelong shen, Wallis, Allen-Zhu, Li, Wang, Wang, and Chen]{hu2022lora}
Edward~J Hu, yelong shen, Phillip Wallis, Zeyuan Allen-Zhu, Yuanzhi Li, Shean Wang, Lu Wang, and Weizhu Chen.
\newblock Lo{RA}: Low-rank adaptation of large language models.
\newblock In \emph{International Conference on Learning Representations}, 2022.

\bibitem[Ilharco et~al.(2021)Ilharco, Wortsman, Wightman, Gordon, Carlini, Taori, Dave, Shankar, Namkoong, Miller, Hajishirzi, Farhadi, and Schmidt]{ilharco_gabriel_2021_5143773}
Gabriel Ilharco, Mitchell Wortsman, Ross Wightman, Cade Gordon, Nicholas Carlini, Rohan Taori, Achal Dave, Vaishaal Shankar, Hongseok Namkoong, John Miller, Hannaneh Hajishirzi, Ali Farhadi, and Ludwig Schmidt.
\newblock Openclip, 2021.
\newblock \url{https://doi.org/10.5281/zenodo.5143773}.

\bibitem[Kamath et~al.(2023)Kamath, Hessel, and Chang]{kamath-etal-2023-whats}
Amita Kamath, Jack Hessel, and Kai-Wei Chang.
\newblock What{'}s {``}up{''} with vision-language models? investigating their struggle with spatial reasoning.
\newblock In \emph{Proceedings of the 2023 Conference on Empirical Methods in Natural Language Processing}, pages 9161--9175, Singapore, 2023. Association for Computational Linguistics.

\bibitem[Karpathy and Fei-Fei(2015)]{karpathy2015deep}
Andrej Karpathy and Li Fei-Fei.
\newblock Deep visual-semantic alignments for generating image descriptions.
\newblock In \emph{Proceedings of the IEEE conference on computer vision and pattern recognition}, pages 3128--3137, 2015.

\bibitem[Krojer et~al.(2022)Krojer, Adlakha, Vineet, Goyal, Ponti, and Reddy]{krojer-etal-2022-imagecode}
Benno Krojer, Vaibhav Adlakha, Vibhav Vineet, Yash Goyal, Edoardo Ponti, and Siva Reddy.
\newblock Image retrieval from contextual descriptions.
\newblock In \emph{Proceedings of the 60th Annual Meeting of the Association for Computational Linguistics (Volume 1: Long Papers)}, pages 3426--3440, Dublin, Ireland, 2022. Association for Computational Linguistics.

\bibitem[Li et~al.(2022{\natexlab{a}})Li, Liu, Li, Zhang, Aneja, Yang, Jin, Hu, Liu, Lee, et~al.]{NEURIPS2022_elevator}
Chunyuan Li, Haotian Liu, Liunian Li, Pengchuan Zhang, Jyoti Aneja, Jianwei Yang, Ping Jin, Houdong Hu, Zicheng Liu, Yong~Jae Lee, et~al.
\newblock Elevater: A benchmark and toolkit for evaluating language-augmented visual models.
\newblock \emph{Advances in Neural Information Processing Systems}, 35:\penalty0 9287--9301, 2022{\natexlab{a}}.

\bibitem[Li et~al.(2022{\natexlab{b}})Li, Liang, Zhao, Cui, Ouyang, Shao, Yu, and Yan]{li2022supervision}
Yangguang Li, Feng Liang, Lichen Zhao, Yufeng Cui, Wanli Ouyang, Jing Shao, Fengwei Yu, and Junjie Yan.
\newblock Supervision exists everywhere: A data efficient contrastive language-image pre-training paradigm.
\newblock In \emph{International Conference on Learning Representations}, 2022{\natexlab{b}}.

\bibitem[Ma et~al.(2023)Ma, Hong, Gul, Gandhi, Gao, and Krishna]{ma2023crepe}
Zixian Ma, Jerry Hong, Mustafa~Omer Gul, Mona Gandhi, Irena Gao, and Ranjay Krishna.
\newblock Crepe: Can vision-language foundation models reason compositionally?
\newblock In \emph{Proceedings of the IEEE/CVF Conference on Computer Vision and Pattern Recognition}, pages 10910--10921, 2023.

\bibitem[Mu et~al.(2022)Mu, Kirillov, Wagner, and Xie]{mu2022slip}
Norman Mu, Alexander Kirillov, David Wagner, and Saining Xie.
\newblock Slip: Self-supervision meets language-image pre-training.
\newblock In \emph{European conference on computer vision}, pages 529--544. Springer, 2022.

\bibitem[Parcalabescu et~al.(2022)Parcalabescu, Cafagna, Muradjan, Frank, Calixto, and Gatt]{parcalabescu2022valse}
Letitia Parcalabescu, Michele Cafagna, Lilitta Muradjan, Anette Frank, Iacer Calixto, and Albert Gatt.
\newblock {VALSE}: A task-independent benchmark for vision and language models centered on linguistic phenomena.
\newblock In \emph{Proceedings of the 60th Annual Meeting of the Association for Computational Linguistics (Volume 1: Long Papers)}, pages 8253--8280, Dublin, Ireland, 2022. Association for Computational Linguistics.

\bibitem[Radenovic et~al.(2023)Radenovic, Dubey, Kadian, Mihaylov, Vandenhende, Patel, Wen, Ramanathan, and Mahajan]{radenovic2023filtering_diht}
Filip Radenovic, Abhimanyu Dubey, Abhishek Kadian, Todor Mihaylov, Simon Vandenhende, Yash Patel, Yi Wen, Vignesh Ramanathan, and Dhruv Mahajan.
\newblock Filtering, distillation, and hard negatives for vision-language pre-training.
\newblock In \emph{Proceedings of the IEEE/CVF conference on computer vision and pattern recognition}, pages 6967--6977, 2023.

\bibitem[Radford et~al.(2021)Radford, Kim, Hallacy, Ramesh, Goh, Agarwal, Sastry, Askell, Mishkin, Clark, et~al.]{radford2021learning}
Alec Radford, Jong~Wook Kim, Chris Hallacy, Aditya Ramesh, Gabriel Goh, Sandhini Agarwal, Girish Sastry, Amanda Askell, Pamela Mishkin, Jack Clark, et~al.
\newblock Learning transferable visual models from natural language supervision.
\newblock In \emph{International conference on machine learning}, pages 8748--8763. PMLR, 2021.

\bibitem[Sahin et~al.(2024)Sahin, Li, Khan, Cremers, and Tresp]{sahin2024enhancing_gmnclip}
Ugur Sahin, Hang Li, Qadeer Khan, Daniel Cremers, and Volker Tresp.
\newblock Enhancing multimodal compositional reasoning of visual language models with generative negative mining.
\newblock In \emph{Proceedings of the IEEE/CVF Winter Conference on Applications of Computer Vision}, pages 5563--5573, 2024.

\bibitem[Singh et~al.(2023)Singh, Zhang, Wang, Wang, Xiong, Du, and Chen]{singh-etal-2023-coarse}
Harman Singh, Pengchuan Zhang, Qifan Wang, Mengjiao Wang, Wenhan Xiong, Jingfei Du, and Yu Chen.
\newblock Coarse-to-fine contrastive learning in image-text-graph space for improved vision-language compositionality.
\newblock In \emph{Proceedings of the 2023 Conference on Empirical Methods in Natural Language Processing}, pages 869--893, Singapore, 2023. Association for Computational Linguistics.

\bibitem[Sun et~al.(2023)Sun, Zhang, Zhang, Shah, Saenko, and Xia]{sun2023dime}
Ximeng Sun, Pengchuan Zhang, Peizhao Zhang, Hardik Shah, Kate Saenko, and Xide Xia.
\newblock Dime-fm: Distilling multimodal and efficient foundation models.
\newblock In \emph{Proceedings of the IEEE/CVF International Conference on Computer Vision}, pages 15521--15533, 2023.

\bibitem[Thrush et~al.(2022)Thrush, Jiang, Bartolo, Singh, Williams, Kiela, and Ross]{thrush2022winoground}
Tristan Thrush, Ryan Jiang, Max Bartolo, Amanpreet Singh, Adina Williams, Douwe Kiela, and Candace Ross.
\newblock Winoground: Probing vision and language models for visio-linguistic compositionality.
\newblock In \emph{Proceedings of the IEEE/CVF Conference on Computer Vision and Pattern Recognition}, pages 5238--5248, 2022.

\bibitem[Tian et~al.(2024)Tian, Fan, Isola, Chang, and Krishnan]{tian2024stablerep}
Yonglong Tian, Lijie Fan, Phillip Isola, Huiwen Chang, and Dilip Krishnan.
\newblock Stablerep: Synthetic images from text-to-image models make strong visual representation learners.
\newblock \emph{Advances in Neural Information Processing Systems}, 36, 2024.

\bibitem[Tong et~al.(2024)Tong, Liu, Zhai, Ma, LeCun, and Xie]{tong2024eyes}
Shengbang Tong, Zhuang Liu, Yuexiang Zhai, Yi Ma, Yann LeCun, and Saining Xie.
\newblock Eyes wide shut? exploring the visual shortcomings of multimodal llms.
\newblock \emph{arXiv preprint arXiv:2401.06209}, 2024.

\bibitem[Vasu et~al.(2023)Vasu, Pouransari, Faghri, Vemulapalli, and Tuzel]{vasu2023mobileclip}
Pavan Kumar~Anasosalu Vasu, Hadi Pouransari, Fartash Faghri, Raviteja Vemulapalli, and Oncel Tuzel.
\newblock Mobileclip: Fast image-text models through multi-modal reinforced training.
\newblock \emph{arXiv preprint arXiv:2311.17049}, 2023.

\bibitem[Wang et~al.(2023)Wang, Lin, Li, Lin, Yang, Zhang, Liu, and Wang]{wang2023equivariant}
Tan Wang, Kevin Lin, Linjie Li, Chung-Ching Lin, Zhengyuan Yang, Hanwang Zhang, Zicheng Liu, and Lijuan Wang.
\newblock Equivariant similarity for vision-language foundation models.
\newblock In \emph{Proceedings of the IEEE/CVF International Conference on Computer Vision}, pages 11998--12008, 2023.

\bibitem[Wortsman et~al.(2022)Wortsman, Ilharco, Kim, Li, Kornblith, Roelofs, Lopes, Hajishirzi, Farhadi, Namkoong, et~al.]{wortsman2022robust}
Mitchell Wortsman, Gabriel Ilharco, Jong~Wook Kim, Mike Li, Simon Kornblith, Rebecca Roelofs, Raphael~Gontijo Lopes, Hannaneh Hajishirzi, Ali Farhadi, Hongseok Namkoong, et~al.
\newblock Robust fine-tuning of zero-shot models.
\newblock In \emph{Proceedings of the IEEE/CVF conference on computer vision and pattern recognition}, pages 7959--7971, 2022.

\bibitem[Wu et~al.(2023)Wu, Peng, Zhou, Xiao, Liu, Yuan, Xuan, Valenzuela, Chen, Wang, et~al.]{wu2023tinyclip}
Kan Wu, Houwen Peng, Zhenghong Zhou, Bin Xiao, Mengchen Liu, Lu Yuan, Hong Xuan, Michael Valenzuela, Xi~Stephen Chen, Xinggang Wang, et~al.
\newblock Tinyclip: Clip distillation via affinity mimicking and weight inheritance.
\newblock In \emph{Proceedings of the IEEE/CVF International Conference on Computer Vision}, pages 21970--21980, 2023.

\bibitem[Yang et~al.(2022)Yang, Li, Zhang, Xiao, Liu, Yuan, and Gao]{yang2022unified_unicl}
Jianwei Yang, Chunyuan Li, Pengchuan Zhang, Bin Xiao, Ce Liu, Lu Yuan, and Jianfeng Gao.
\newblock Unified contrastive learning in image-text-label space.
\newblock In \emph{Proceedings of the IEEE/CVF Conference on Computer Vision and Pattern Recognition}, pages 19163--19173, 2022.

\bibitem[Yang et~al.(2023)Yang, Deng, An, Li, Feng, Guo, Yang, and Liu]{yang2023alip}
Kaicheng Yang, Jiankang Deng, Xiang An, Jiawei Li, Ziyong Feng, Jia Guo, Jing Yang, and Tongliang Liu.
\newblock Alip: Adaptive language-image pre-training with synthetic caption.
\newblock In \emph{Proceedings of the IEEE/CVF International Conference on Computer Vision}, pages 2922--2931, 2023.

\bibitem[Yuksekgonul et~al.(2023)Yuksekgonul, Bianchi, Kalluri, Jurafsky, and Zou]{yuksekgonul2023when}
Mert Yuksekgonul, Federico Bianchi, Pratyusha Kalluri, Dan Jurafsky, and James Zou.
\newblock When and why vision-language models behave like bags-of-words, and what to do about it?
\newblock In \emph{The Eleventh International Conference on Learning Representations}, 2023.

\bibitem[Zhai et~al.(2023)Zhai, Mustafa, Kolesnikov, and Beyer]{zhai2023sigmoid}
Xiaohua Zhai, Basil Mustafa, Alexander Kolesnikov, and Lucas Beyer.
\newblock Sigmoid loss for language image pre-training.
\newblock In \emph{Proceedings of the IEEE/CVF International Conference on Computer Vision}, pages 11975--11986, 2023.

\bibitem[Zhang et~al.(2023)Zhang, Awal, and Agrawal]{zhang2023contrasting}
Le Zhang, Rabiul Awal, and Aishwarya Agrawal.
\newblock Contrasting intra-modal and ranking cross-modal hard negatives to enhance visio-linguistic fine-grained understanding.
\newblock \emph{arXiv preprint arXiv:2306.08832}, 2023.

\bibitem[Zhao et~al.(2022)Zhao, Zhang, Zhu, Shen, Lee, Lu, and Yin]{zhao2022vl}
Tiancheng Zhao, Tianqi Zhang, Mingwei Zhu, Haozhan Shen, Kyusong Lee, Xiaopeng Lu, and Jianwei Yin.
\newblock Vl-checklist: Evaluating pre-trained vision-language models with objects, attributes and relations.
\newblock \emph{arXiv preprint arXiv:2207.00221}, 2022.

\end{thebibliography}
}

\clearpage

\appendix
\onecolumn

\begin{center}
    {\Large \bf
        Exploring the Spectrum of Visio-Linguistic Compositionality and Recognition
    } {\vskip 2mm} 
    {\Large
        Supplementary Material
    }
\end{center}
\vspace{5mm}
In this supplementary material, we offer further details. First, \cref{sec:sup_eval_details} contains information about the specific compositionality benchmarks and models utilized within our evaluation framework. Specifically, \cref{tab:sup_benchmarks} provides a comprehensive list of compositionality benchmarks, organized into I2T, T2I, and Group categories, detailing their image sources and associated tasks. In addition, \cref{tab:sup_pretrained,tab:sup_finetuned} detail the lists of pre-trained and fine-tuned CLIP models utilized in our analysis. 

We then expand the observations made in the main paper with a more detailed examination at the individual benchmark level in~\cref{sec:sup_detailed_res}. Mirroring the~\cref{fig:teaser}, \cref{fig:sup_1_benchmark} offers a detailed exploration of the trade-offs between compositionality and recognition across VLMs, specifically focusing on individual compositionality benchmarks. Similarly, we also detail the trajectories of fine-tuning models across compositionality and various recognition tasks (e.g., zero-shot classification and retrieval) from~\cref{fig:sup_2_wiseft,fig:sup_3_wiseft,fig:sup_4_wiseft}. Lastly, we provide comprehensive numerical results for the fine-tuned models, including additional models not featured within the figures, in~\cref{tab:sup_report_finetuned}. 

\section{Evaluation Toolkit Details}
\label{sec:sup_eval_details}

\noindent
\begin{table*}[ht]
\centering
\resizebox{.99\linewidth}{!}{%
    \begin{tabular}{p{3.9cm} p{5.2cm} p{10cm}}
    \toprule
        Benchmark  &  Image source & Tasks and Subtasks \\
    \midrule
    ARO~\cite{yuksekgonul2023when} & COCO, Visual Genome, Flickr30k & VG\_Relation, VG\_Attribution, Flickr30k\_Order, COCO\_Order \\
    CREPE (Productivity)~\cite{ma2023crepe} & Visual Genome & Atomic Foils, Negate, Swap \\
    SugarCrepe~\cite{hsieh2024sugarcrepe} & COCO & Add\_\{object, attribute\}, Replace\_\{object, attribute, relation\}, Swap\_\{object, attribute\}\\
    VALSE~\cite{parcalabescu2022valse}  &  Visual7w, COCO, SWiG, VisDial\_v1.0, FOIL-it & Actions\_\{swap, replacement\}, Coreference\_\{hard, standard\}, Counting\_\{adversarial, hard, small\}, Existence, Foil-it, Plurals, Relations \\
    VL-Checklist~\cite{zhao2022vl} &  Visual Genome, SWiG, COCO, HAKE, HICO\_Det, Pic, HCVRD, OpenImages & Object\_Location\_\{center, margin, mid\}, Object\_Size\_\{large, medium, small\}, Attribute\_\{action, color, material, size, state\}, Relation\_\{action, spatial\} \\
    WhatsUp~\cite{kamath-etal-2023-whats} & Controlled\_Images (\textit{self-captured}), COCO, GQA & Controlled\_Images\_\{A, B\}, COCO\_QA\_\{One, Two\}, VG\_QA\_\{One, Two\} \\
    \cmidrule(lr){1-3}
    ImageCoDe~\cite{krojer-etal-2022-imagecode} & OpenImages, MSRVTT, Video-Storytelling, YouCook & Static (\eg, images), Video (\eg, videos) \\
    SVO Probes~\cite{hendricks-nematzadeh-2021-probing} & Google Image Search API & Subject, Verb, Object \\
    \cmidrule(lr){1-3}
    Winoground~\cite{thrush2022winoground} & Getty Images & - \\ 
    ColorSwap~\cite{burapacheep2024colorswap} & Generative models (\eg, Midjourney, DALLE3, and StableDiffusion) & - \\
    EqBen~\cite{wang2023equivariant} & Action Genome (AG), GEBC, YouCook2, Kubric, StableDiffusion (SD)  & EQ-AG, EQ-GEBC, EQ-YouCook2, EQ-Kubric\_\{location, counting, attribute\}, EQ-SD\\
    MMVP-VLM~\cite{tong2024eyes} & - & Color and Appearance, Orientation and Direction, Positional and Relational Context, Presence of Specific Features, Quantity and Count, State and Condition, Structural Characteristics, Texts, Viewpoint and Perspective \\
    \midrule
    Total & 12 & \\
    \bottomrule
\end{tabular}
}
\caption{A complete list of compositionality benchmarks implemented in our evaluation framework. 
In the table, benchmarks are organized into Image-to-Text (I2T), Text-to-Image (T2I), and Group settings, distinguished by horizontal lines, as exemplified in the main paper.
For evaluation, subtasks of a task, identified by enclosed brackets, are aggregated to obtain individual task performance. The overall evaluation metric is then derived by averaging these task-specific performances. While we employ unweighted averaging for aggregation, for SugarCrepe~\cite{hsieh2024sugarcrepe} and ImageCoDe~\cite{krojer-etal-2022-imagecode}, we utilize weighted averaging by sample numbers, in alignment with their official implementations.
}
\label{tab:sup_benchmarks}
\vspace{-2mm}
\end{table*}

\clearpage

\begin{table*}[ht]
\centering
\resizebox{.99\linewidth}{!}{%
    \begin{tabular}{p{2.2cm} >{\centering\arraybackslash}p{0.8cm} p{16cm}}
    \toprule  Family  &  Count & Models \\
    \midrule
    OpenCLIP~\cite{ilharco_gabriel_2021_5143773} &    85    & {\footnotesize yfcc15m:RN50, yfcc15m:RN101, cc12m:RN50, openai:RN50, openai:RN101, openai:RN50x4, openai:RN50x16, openai:RN50x64, openai:ViT-B-32, openai:ViT-B-16, openai:ViT-L-14, metaclip\_400m:ViT-B-32-quickgelu, metaclip\_400m:ViT-B-16-quickgelu, metaclip\_400m:ViT-L-14-quickgelu, metaclip\_fullcc:ViT-B-32-quickgelu, metaclip\_fullcc:ViT-B-16-quickgelu, metaclip\_fullcc:ViT-L-14-quickgelu, metaclip\_fullcc:ViT-H-14-quickgelu, datacomp\_s\_s13m\_b4k:ViT-B-32, datacomp\_m\_s128m\_b4k:ViT-B-32, datacomp\_xl\_s13b\_b90k:ViT-B-32, datacomp\_xl\_s13b\_b90k:ViT-B-16, datacomp\_l\_s1b\_b8k:ViT-B-16, laion2b\_s34b\_b79k:ViT-B-32, laion2b\_s34b\_b88k:ViT-B-16, laion2b\_s32b\_b82k:ViT-L-14, laion2b\_s34b\_b88k:ViT-g-14, laion2b\_s32b\_b79k:ViT-H-14, laion2b\_s39b\_b160k:ViT-bigG-14, laion400m\_s11b\_b41k:EVA01-g-14, merged2b\_s8b\_b131k:EVA02-B-16, merged2b\_s4b\_b131k:EVA02-L-14, laion2b\_s4b\_b115k:EVA02-E-14, datacomp1b:ViT-L-14-CLIPA, datacomp1b:ViT-H-14-CLIPA, datacomp1b:ViT-bigG-14-CLIPA, dfn2b:ViT-B-16, dfn2b:ViT-L-14-quickgelu, dfn5b:ViT-H-14-quickgelu, webli:ViT-B-16-SigLIP, webli:ViT-L-16-SigLIP-256, webli:ViT-SO400M-14-SigLIP, laion400m\_s13b\_b51k:convnext\_btase, laion2b\_s26b\_b102k\_augreg:convnext\_large\_d, laion2b\_s34b\_b82k\_augreg:convnext\_xxlarge, laion2b\_s13b\_b90k:coca\_ViT-B-32, laion2b\_s13b\_b90k:coca\_ViT-L-14, laion2b\_s12b\_b32k:roberta-ViT-B-32, laion5b\_s13b\_b90k:xlm-roberta-base-ViT-B-32, yfcc15m:RN50-quickgelu, cc12m:RN50-quickgelu, openai:RN101-quickgelu, yfcc15m:RN101-quickgelu, laion400m\_e32:ViT-B-32, commonpool\_m\_clip\_s128m\_b4k:ViT-B-32, commonpool\_m\_laion\_s128m\_b4k:ViT-B-32, commonpool\_m\_image\_s128m\_b4k:ViT-B-32, commonpool\_m\_text\_s128m\_b4k:ViT-B-32, commonpool\_m\_basic\_s128m\_b4k:ViT-B-32, commonpool\_m\_s128m\_b4k:ViT-B-32, commonpool\_s\_clip\_s13m\_b4k:ViT-B-32, commonpool\_s\_laion\_s13m\_b4k:ViT-B-32, commonpool\_s\_image\_s13m\_b4k:ViT-B-32, commonpool\_s\_text\_s13m\_b4k:ViT-B-32, commonpool\_s\_basic\_s13m\_b4k:ViT-B-32, commonpool\_s\_s13m\_b4k:ViT-B-32, datacomp\_s34b\_b86k:ViT-B-32-256, laion400m\_e32:ViT-B-32-quickgelu, laion400m\_e32:ViT-B-16, commonpool\_l\_clip\_s1b\_b8k:ViT-B-16, commonpool\_l\_laion\_s1b\_b8k:ViT-B-16, commonpool\_l\_image\_s1b\_b8k:ViT-B-16, commonpool\_l\_text\_s1b\_b8k:ViT-B-16, commonpool\_l\_basic\_s1b\_b8k:ViT-B-16, commonpool\_l\_s1b\_b8k:ViT-B-16, laion400m\_e32:ViT-B-16-plus-240, commonpool\_xl\_clip\_s13b\_b90k:ViT-L-14, commonpool\_xl\_laion\_s13b\_b90k:ViT-L-14, commonpool\_xl\_s13b\_b90k:ViT-L-14, laion400m\_e32:ViT-L-14, laion2b\_s13b\_b82k:convnext\_base\_w, laion2b\_s13b\_b82k\_augreg:convnext\_base\_w, laion\_aesthetic\_s13b\_b82k:convnext\_base\_w, laion\_aesthetic\_s13b\_b82k:convnext\_base\_w\_320, webli:ViT-B-16-SigLIP-384} \\
    
    \cmidrule(lr){1-3}

    SLIP~\cite{mu2022slip} &    10    &  {\footnotesize yfcc15m:ViT-S-CLIP, yfcc15m:ViT-S-SLIP, yfcc15m:ViT-B-CLIP, yfcc15m:ViT-B-SLIP, yfcc15m:ViT-L-CLIP, yfcc15m:ViT-L-SLIP, cc3m:ViT-B-CLIP, cc3m:ViT-B-SLIP, cc12m:ViT-B-CLIP, cc12m:ViT-B-SLIP} \\
    
    CyCLIP~\cite{goel2022cyclip} &    4     &  {\footnotesize cc3m:CLIP, cc3m:CyCLIP, cc3m:i-CyCLIP, cc3m:c-CyCLIP} \\
    
    MERU~\cite{desai2023meru} &    6     &   {\footnotesize redcaps:CLIP-ViT-S, redcaps:MERU-ViT-S, redcaps:CLIP-ViT-B, redcaps:MERU-ViT-B, redcaps:CLIP-ViT-L, redcaps:MERU-ViT-L} \\
    
    DeCLIP~\cite{li2022supervision,cui2022democratizing_declip} &    9     &  {\footnotesize yfcc15m:CLIP\_RN50, yfcc15m:DeCLIP\_RN50, yfcc15m:CLIP\_ViT-B-32, yfcc15m:SLIP\_ViT-B-32, yfcc15m:FILIP\_ViT-B-32, yfcc15m:DeFILIP\_ViT-B-32, yfcc15m:DeCLIP\_ViT-B-32, declip88m:DeCLIP\_RN50, declip88m:DeCLIP\_ViT-B-32} \\
    
    UniCL~\cite{yang2022unified_unicl} &    5     &  {\footnotesize yfcc14m:swin\_tiny, in21k\_yfcc14m:swin\_tiny, yfcc14m:swin\_base, in21k\_yfcc14m:swin\_base, in21k\_yfcc14m\_gcc15m:swin\_base} \\
    
    DiHT~\cite{radenovic2023filtering_diht} &    4     &  {\footnotesize laion2b:diht\_vitb16\_224px, laion2b:diht\_vitb32\_224px, laion2b:diht\_vitl14\_224px, laion2b:diht\_vitl14\_336px} \\
    
    MobileCLIP~\cite{vasu2023mobileclip} &    5     &   {\footnotesize datacomp-dr:MobileCLIP-S0, datacomp-dr:MobileCLIP-S1, datacomp-dr:MobileCLIP-S2, datacomp-dr:MobileCLIP-B, datacomp-dr:MobileCLIP-B-LT} \\
    
    TinyCLIP~\cite{wu2023tinyclip} &    9     &  {\footnotesize yfcc15m:ViT-8M-16-Text-3M, yfcc15m:ViT-39M-16-Text-19M, laion400m:ViT-40M-32-Text-19M, laion400m:ViT-61M-32-Text-29M, laion400m:auto-ViT-22M-32-Text-10M, laion400m:auto-ViT-45M-32-Text-18M, laion400m:auto-ViT-63M-32-Text-31M, laion-yfcc:auto-ViT-45M-32-Text-18M, laion-yfcc:auto-ViT-63M-32-Text-31M} \\
    
    DIME-FM~\cite{sun2023dime} &    2     &  {\footnotesize in21k\_yfcc14m\_gcc15m:NLP-ViT-B-32, in21k\_yfcc14m\_gcc15m:Prompts-ViT-B-32} \\
    
    
    ALIP~\cite{yang2023alip} &    1     & {\footnotesize yfcc15m:ALIP} \\
    
    LaCLIP~\cite{fan2024improving_laclip} &    12    &  {\footnotesize cc3m:CLIP\_ViT-B-16, cc3m:LaCLIP\_ViT-B-16, cc12m:CLIP\_ViT-B-16, cc12m:LaCLIP\_ViT-B-16, redcaps:CLIP\_ViT-B-16, redcaps:LaCLIP\_ViT-B-16, laion400m:CLIP\_ViT-B-32, laion400m:LaCLIP\_ViT-B-32, laion400m:CLIP\_ViT-B-16, laion400m:LaCLIP\_ViT-B-16, laion400m:CLIP\_ViT-L-14, laion400m:LaCLIP\_ViT-L-14} \\
    
    SynthCLIP~\cite{hammoud2024synthclip} &    4     & {\footnotesize synthci10m:CLIP\_ViT-B-16, synthci20m:CLIP\_ViT-B-16, synthci30m:CLIP\_ViT-B-16, cc12m:CLIP\_ViT-B-16} \\
    
    StableRep~\cite{tian2024stablerep} &    8     &  {\footnotesize laion3m:CLIP\_vitb16, laion3m:StableRep-pp\_vitb16, laion10m:CLIP\_vitb16, laion10m:StableRep-pp\_vitb16, laion20m:CLIP\_vitb16, laion20m:StableRep-pp\_vitb16, laion50m:CLIP\_vitb16, laion50m:StableRep-pp\_vitb16} \\
    
    Scaling~\cite{fan2023scaling} &    30    &  {\footnotesize syn1m:CLIP\_ViT-B-16, real1m:CLIP\_ViT-B-16, synreal1m:CLIP\_ViT-B-16, syn2m:CLIP\_ViT-B-16, real2m:CLIP\_ViT-B-16, synreal2m:CLIP\_ViT-B-16, syn4m:CLIP\_ViT-B-16, real4m:CLIP\_ViT-B-16, synreal4m:CLIP\_ViT-B-16, syn8m:CLIP\_ViT-B-16, real8m:CLIP\_ViT-B-16, synreal8m:CLIP\_ViT-B-16, syn16m:CLIP\_ViT-B-16, real16m:CLIP\_ViT-B-16, synreal16m:CLIP\_ViT-B-16, syn32m:CLIP\_ViT-B-16, real32m:CLIP\_ViT-B-16, synreal32m:CLIP\_ViT-B-16, syn64m:CLIP\_ViT-B-16, real64m:CLIP\_ViT-B-16, synreal64m:CLIP\_ViT-B-16, syn128m:CLIP\_ViT-B-16, real128m:CLIP\_ViT-B-16, synreal128m:CLIP\_ViT-B-16, syn256m:CLIP\_ViT-B-16, real256m:CLIP\_ViT-B-16, synreal256m:CLIP\_ViT-B-16, syn371m:CLIP\_ViT-B-16, real371m:CLIP\_ViT-B-16, synreal371m:CLIP\_ViT-B-16} \\
    
    \midrule
    Total & 194 & \\
    \bottomrule
\end{tabular}
}
\caption{A complete list of the pre-trained CLIP models and their respective architectures specifying pre-trained data, formatted as \texttt{data}:\texttt{architecture} in our study. For the OpenCLIP model, we directly load it, while for the others, we have acquired the corresponding checkpoints from each official repository. 
}
\label{tab:sup_pretrained}
\vspace{-2mm}
\end{table*}

\begin{table*}[ht]
\centering
\resizebox{.99\linewidth}{!}{%
    \begin{tabular}{p{2.5cm} >{\centering\arraybackslash}p{0.8cm} p{14cm}}
    \toprule  Family  &  Count & Models \\
    \midrule
    NegCLIP~\cite{yuksekgonul2023when} & 1 &  coco-ft:NegCLIP \\
    CE-CLIP~\cite{zhang2023contrasting} & 1 & coco-ft:CE-CLIP \\
    GNM-CLIP~\cite{sahin2024enhancing_gmnclip} & 1 & coco-ft:GNM-CLIP \\
    TSVLC~\cite{doveh2023teaching} & 2 & cc3m-ft:TSVLC-Negs\_RB, cc3m-ft:TSVLC-Negs\_LLM \\ 
    DAC~\cite{doveh2023dense} & 2 & cc3m-ft:DAC-LLM, cc3m-ft:DAC-SAM \\
    CLoVe~\cite{castro2024clove} & 1 & laioncoco600m-ft:CLoVe \\
    \midrule
    Total & 8 &  \\
    \bottomrule
\end{tabular}
}
\caption{A complete list of the fine-tuning methodologies of CLIP and their respective architectures specifying fine-tuning data, formatted as \texttt{data}:\texttt{architecture} in our study. We have obtained the corresponding checkpoints from the official repository and implemented an evaluation pipeline aligned to each repository. A major adjustment involves applying \texttt{quick\_gelu=True} when loading models via \textit{open\_clip}~\cite{ilharco_gabriel_2021_5143773} with the fine-tuned checkpoints. This resolves consistency issues across pre-/fine-tuning and evaluation stage, especially affecting NegCLIP~\cite{yuksekgonul2023when}, CE-CLIP~\cite{zhang2023contrasting}, and GNM-CLIP~\cite{sahin2024enhancing_gmnclip}. As note, we interpolate the model parameters of each fine-tuned model with the pre-trained ViT-B/32~\cite{radford2021learning} when applying WiSE-FT~\cite{wortsman2022robust}, adjusting the interpolation weight $\alpha$ from 0 to 1 in steps of 0.1. This yields $8 \times 9 = 72$ intermediate models, resulting in total 274 models for our study, including 194 pre-trained, 8 fine-tuned, and the 72 intermediate models. As TSVLC~\cite{doveh2023teaching} and DAC~\cite{doveh2023dense} applies LoRA~\cite{hu2022lora} for fine-tuning, we only interpolate the model weights corresponding to LoRA layer, maintaining the original pre-trained weights from CLIP ViT-B/32.
}
\label{tab:sup_finetuned}
\vspace{-2mm}
\end{table*}

\section{Additional Benchmark-level Analysis}
\label{sec:sup_detailed_res}
While our study primarily focused on the average characteristics of compositionality benchmarks in conjunction with recognition abilities, we now dissect these at the individual benchmark level for a more granular analysis. 

\subsection{A Holistic View Between Compositionality and Recognition}
In~\cref{fig:sup_1_benchmark}, we break down the overall compositionality performance presented in~\cref{fig:teaser} into individual benchmarks, presenting a total of 12. Aligned to the observations made in the main paper, pre-trained CLIP models exhibit positive correlations between compositionality and recognition, whereas models fine-tuned specifically for compositionality demonstrate trade-offs between these two aspects in general.

In the context of fine-tuning, our benchmark-level analysis reveals varied outcomes: while fine-tuning significantly enhances compositionality in some cases, but in others cases, fine-tuned models can show minimal improvements in compositionality. Specifically, benchmarks such as ARO, CREPE, SugarCrepe, VALSE, and VL-Checklist demonstrate that fine-tuning can effectively boost compositionality at the cost of recognition performance. Conversely, benchmarks like ImageCoDe, SVO Probes, ColorSwap, EqBen, and MMVP-VLM exhibit little to no significant benefit from fine-tuning, aligning closely with the performance trajectory of pre-trained models. Notably, for Winoground, fine-tuning not only fails to enhance but may even diminish both recognition and compositionality. 
Another notable finding is the lack of correlation between compositionality and recognition in the WhatsUp benchmark. Exploring these instances of underperformance in context of both pre-training and fine-tuning approaches presents an intriguing research avenue.

\subsection{Fine-tuning Trajectories of Pre-trained CLIP}
\begin{figure*}[t]
  \centering
   \includegraphics[width=0.95\linewidth]{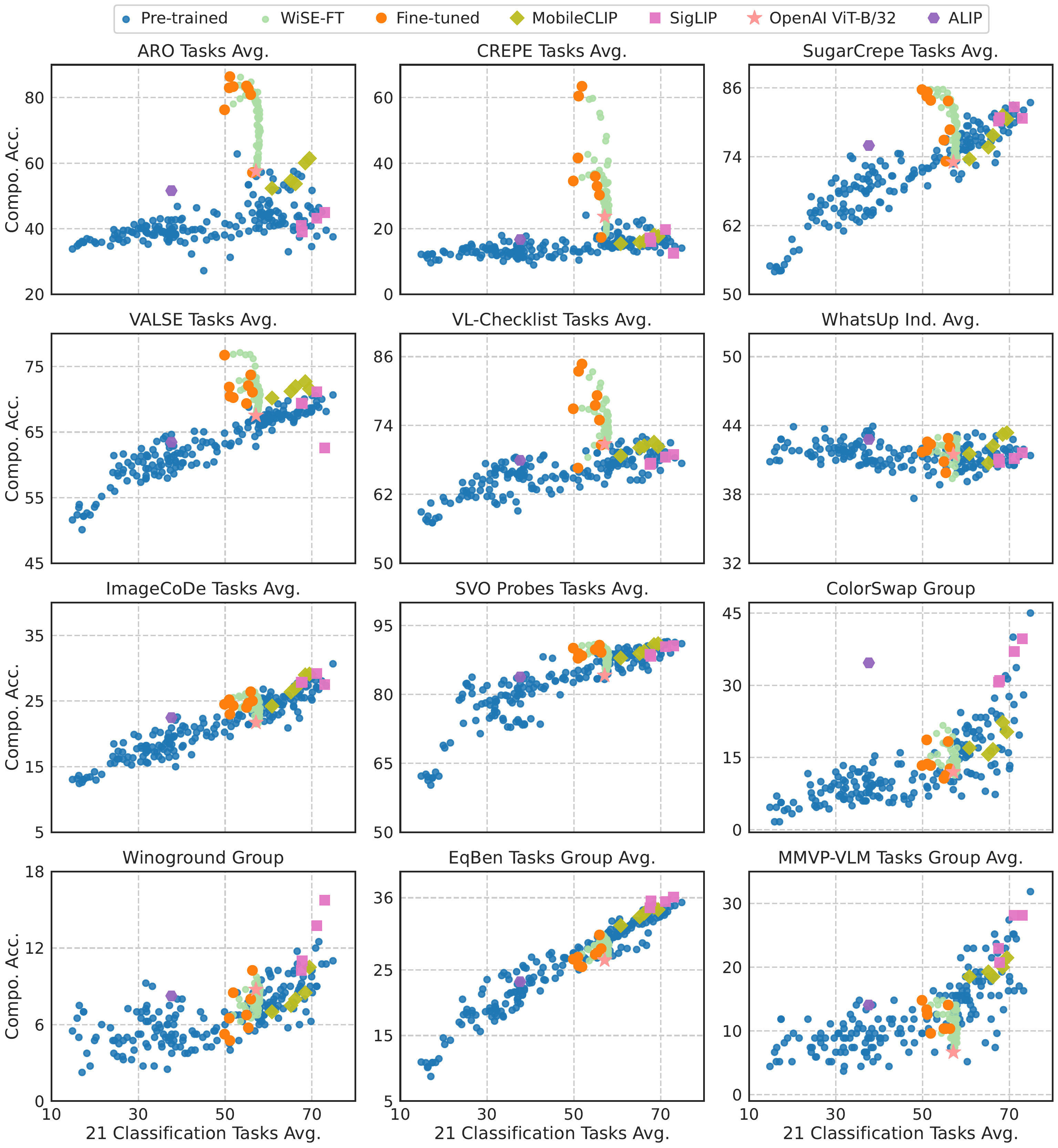}
   \vspace{-2mm}
   \caption{A comparative analysis of pre-trained and fine-tuned CLIP models in both compositionality and recognition at a benchmark-level. Pre-trained models generally show a positive correlation between compositionality and zero-shot classification with the exception of WhatsUp benchmark. In contrast, fine-tuned models exhibit mixed properties across different benchmarks. 
   }
   \label{fig:sup_1_benchmark}
   \vspace{-2mm}
\end{figure*}

\begin{figure*}[t]
  \centering
   \includegraphics[width=0.95\linewidth]{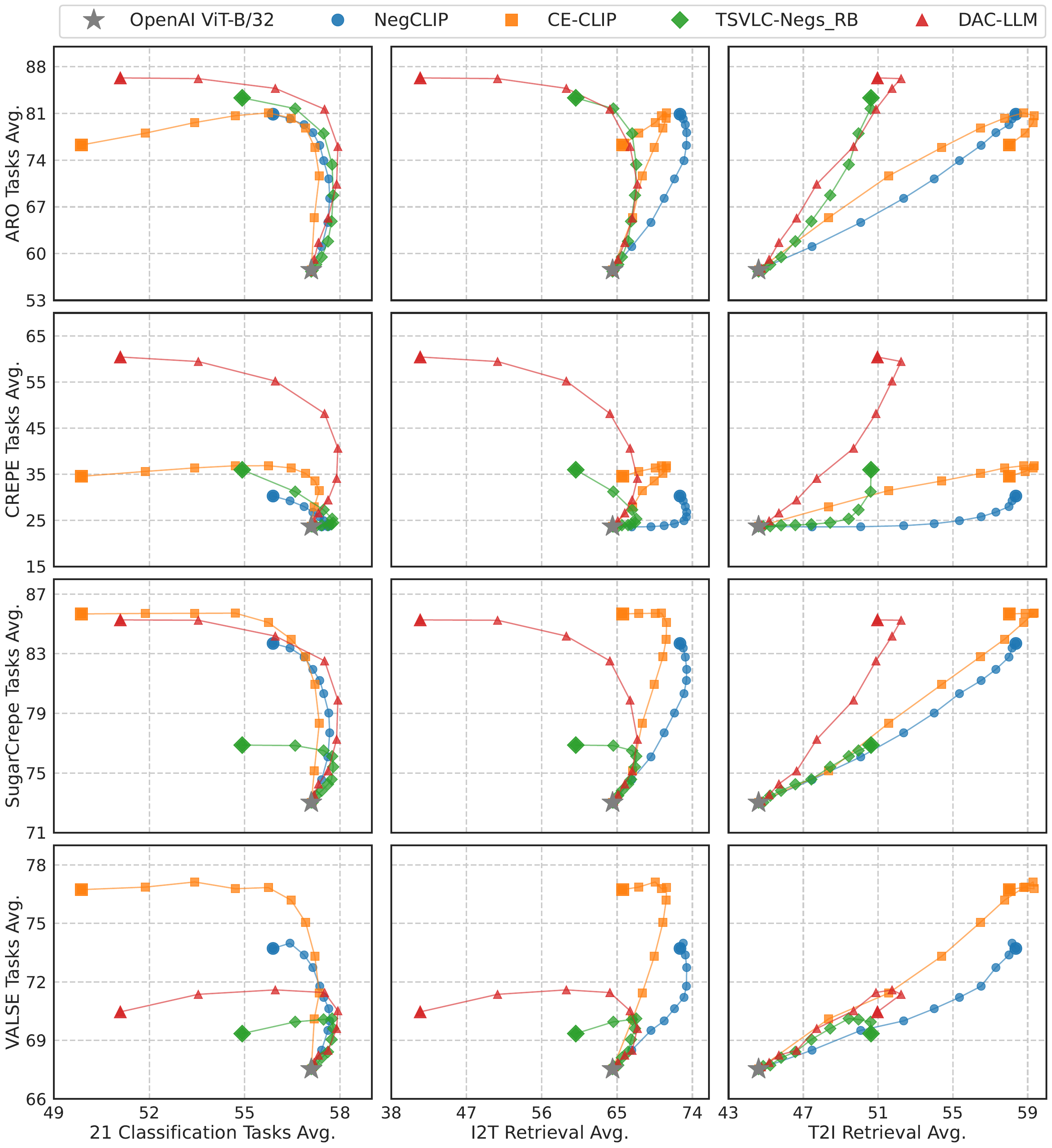}
   \vspace{-2mm}
   \caption{Trade-offs between compositionality and recognition tasks for fine-tuned models across each individual compositionality benchmark. It includes the ARO, CREPE, SugarCrepe, and VALSE datasets for each row, while recognition is evaluated through zero-shot classification, image-to-text (I2T) retrieval, and text-to-image (T2I) retrieval tasks for each column.
   }
   \label{fig:sup_2_wiseft}
   \vspace{-2mm}
\end{figure*}

\begin{figure*}[t]
  \centering
   \includegraphics[width=0.95\linewidth]{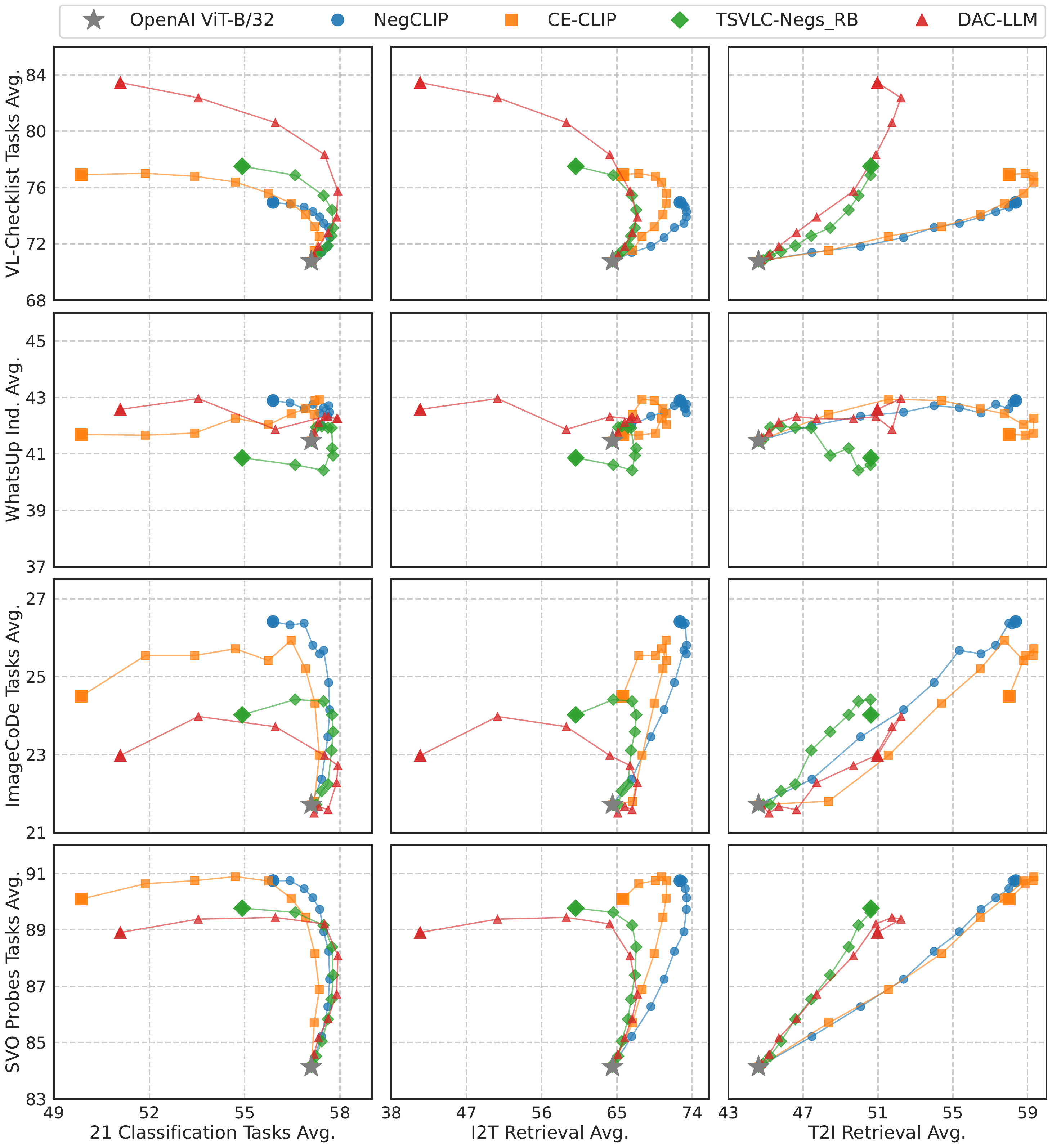}
   \vspace{-2mm}
   \caption{Trade-offs between compositionality and recognition tasks for fine-tuned models across each individual compositionality benchmark. It includes the VL-Checklist, WhatsUp, ImageCoDe, and SVO Probes datasets for each row, while recognition is evaluated through zero-shot classification, image-to-text (I2T) retrieval, and text-to-image (T2I) retrieval tasks for each column.
   }
   \label{fig:sup_3_wiseft}
   \vspace{-2mm}
\end{figure*}

\begin{figure*}[t]
  \centering
   \includegraphics[width=0.95\linewidth]{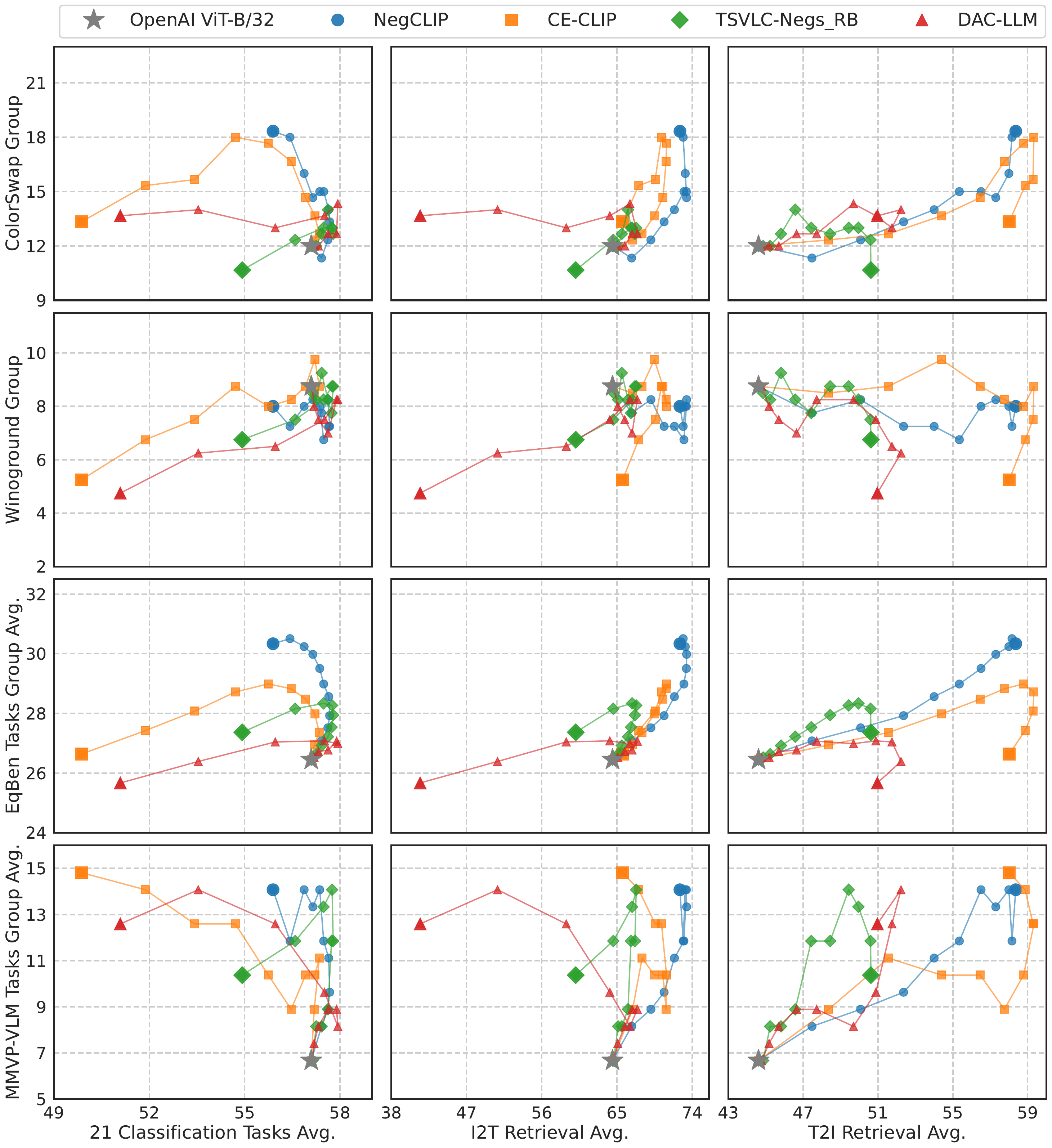}
   \vspace{-2mm}
   \caption{Trade-offs between compositionality and recognition tasks for fine-tuned models across each individual compositionality benchmark. It includes the ColorSwap, Winoground, EqBen, and MMVP-VLM datasets for each row, while recognition is evaluated through zero-shot classification, image-to-text (I2T) retrieval, and text-to-image (T2I) retrieval tasks for each column.
   }
   \label{fig:sup_4_wiseft}
   \vspace{-2mm}
\end{figure*}

We also present the detailed trade-offs between compositionality and recognition via WiSE-FT~\cite{wortsman2022robust}, as illustrated from~\cref{fig:sup_2_wiseft,fig:sup_3_wiseft,fig:sup_4_wiseft}~covering 12 benchmarks. Consistent with the findings presented in the main paper, specific observations at the benchmark level reveal that: (1) compositionality comes at the expense of zero-shot classification accuracy; (2) the effectiveness of image-to-text (I2T) retrieval tasks is influenced by the dataset utilized for fine-tuning; and (3) fine-tuning for compositionality also brings text-to-image (T2I) retrieval performances. Furthermore, in line with the observation from~\cref{fig:sup_1_benchmark}, it is evident that fine-tuning does not significantly improve compositionality on the WhatsUp and Winoground benchmarks.

\clearpage

\clearpage

\subsection{Benchmarking Fine-tuning Methods}
In~\cref{tab:sup_report_finetuned}, we showcase comprehensive benchmarking results for fine-tuning methods aimed at improving compositionality, which are publicly released at the time of our submission. We aimed for reproducibility and fairness in our comparisons, by evaluating on consistent benchmarks and using the same metric. We report the performances across 12 compositionality benchmarks and also zero-shot recognition tasks, including classification and retrieval. We note a decrease in classification accuracy among all fine-tuned models, with the top-scoring models differing based on the specific task at hand.

\newcommand*\rot{\rotatebox{90}}
\begin{table*}[ht]
\centering
\resizebox{.96\linewidth}{!}{%
    \begin{tabular}{rcccccccccccccccccccc}
    \toprule
    Model &  \rot{ARO~\cite{yuksekgonul2023when}}  &  \rot{CREPE~\cite{ma2023crepe}}  &  \rot{SugarCrepe~\cite{hsieh2024sugarcrepe}}  &  \rot{VALSE~\cite{parcalabescu2022valse}}  &  \rot{VL-Checklist~\cite{zhao2022vl}}  &  \rot{WhatsUp~\cite{kamath-etal-2023-whats}}  &  \rot{ImageCoDe~\cite{krojer-etal-2022-imagecode}}  &  \rot{SVO Probes~\cite{hendricks-nematzadeh-2021-probing}}  &  \rot{ColorSwap~\cite{burapacheep2024colorswap}}  &  \rot{Winoground~\cite{thrush2022winoground}}  &  \rot{EqBen~\cite{wang2023equivariant}}  &  \rot{MMVP-VLM~\cite{tong2024eyes}}  &  \rot{COCO I2T Retrieval}  &  \rot{COCO T2I Retrieval}  &  \rot{Flickr30k I2T Retrieval~}  &  \rot{Flickr30k T2I Retrieval~}  &  \rot{21 Classification Avg.~}  &  \rot{I2T Retrieval Avg.}  &  \rot{T2I Retrieval Avg.}  &  \rot{Comp. Avg.}  \\
    \midrule
    OpenAI ViT-B/32~\cite{radford2021learning} & 57.6  &  23.7   &     73.0     &  67.5   &      70.8      &   41.5    &    21.7     &     84.1     &    12.0     &     8.8      &  26.5   &    6.7     &    50.1    &    30.5    &     78.8     &     58.8     &      \textbf{57.1}      &   64.5    &   44.6    &    41.2    \\
    \cmidrule(lr){1-1} \cmidrule(lr){2-13} \cmidrule(lr){14-17} \cmidrule(lr){18-21}
    NegCLIP~\cite{yuksekgonul2023when} & 80.9  &  30.3   &     83.7     &  73.7   &      75.0      &   \textbf{42.9}    &    \textbf{26.4}     &     \textbf{90.7}     &    18.3     &     8.0      &  \textbf{30.3}   &    14.1    &    \textbf{59.3}    &    45.2    &     \textbf{85.7}     &     \textbf{71.6}     &      55.9      &   \textbf{72.5}    &   \textbf{58.4}    &    47.9    \\
    
    CE-CLIP~\cite{zhang2023contrasting} & 76.3  &  34.6   &     \textbf{85.7}     &  \textbf{76.7}   &      76.9      &   41.7    &    24.5     &     90.1     &    13.3     &     5.2      &  26.6   &    \textbf{14.8}    &    56.0    &    \textbf{47.1}    &     75.3     &     68.9     &      49.9      &   65.7    &   58.0    &    47.2    \\
    
    GNM-CLIP~\cite{sahin2024enhancing_gmnclip} & 57.1  &  17.3   &     78.7     &  71.1   &      70.6      &   42.1    &    25.0     &     89.2     &    12.7     &     \textbf{10.2}     &  28.2   &    10.4    &    58.1    &    41.1    &     82.9     &     68.8     &      56.3      &   70.5    &   54.9    &    42.7    \\
    \cmidrule(lr){1-1} \cmidrule(lr){2-13} \cmidrule(lr){14-17} \cmidrule(lr){18-21}
    TSVLC\_RB~\cite{doveh2023teaching} & 83.5  &  36.0   &     76.9     &  69.4   &      77.5      &   40.9    &    24.0     &     89.8     &    10.7     &     6.8      &  27.4   &    10.4    &    46.1    &    36.4    &     74.0     &     64.9     &      54.9      &   60.1    &   50.6    &    46.1    \\
    
    TSVLC\_RB\_LLM~\cite{doveh2023teaching} & 82.7  &  33.0   &     73.2     &  72.1   &      79.2      &   39.9    &    24.7     &     89.7     &    11.3     &     5.8      &  27.6   &    10.4    &    46.4    &    36.6    &     74.9     &     65.1     &      55.3      &   60.7    &   50.9    &    45.8    \\
    
    DAC-LLM~\cite{doveh2023dense} & \textbf{86.4}  &  60.4   &     85.3     &  70.5   &      83.5      &   42.6    &    23.0     &     88.9     &    13.7     &     4.8      &  25.7   &    12.6    &    29.9    &    37.3    &     53.0     &     64.6     &      51.1      &   41.5    &   51.0    &    \textbf{49.8}    \\
    
    DAC-SAM~\cite{doveh2023dense} & 83.3  &  \textbf{63.4}   &     83.8     &  70.3   &      \textbf{84.7}      &   42.4    &    24.3     &     88.4     &    13.3     &     8.5      &  25.5   &    9.6     &    33.1    &    33.9    &     59.7     &     61.7     &      51.9      &   46.4    &   47.8    &    \textbf{49.8}    \\
    \cmidrule(lr){1-1} \cmidrule(lr){2-13} \cmidrule(lr){14-17} \cmidrule(lr){18-21}
    CLoVe~\cite{castro2024clove} & 83.0  &  41.6   &     84.5     &  71.9   &      66.6      &   41.8    &    25.2     &     87.9     &    \textbf{18.7}     &     6.5      &  27.0   &    13.3    &    48.2    &    42.7    &     69.5     &     68.7     &      50.9      &   58.9    &   55.7    &    47.3    \\
    \bottomrule
    \end{tabular}
}
\caption{Benchmarking fine-tuning methods, reporting both compositionality and recognition tasks. Compared to the pre-trained OpenAI model, fine-tuning results in decreased classification performance, and there is no single model that outperforms in all tasks.
}
\label{tab:sup_report_finetuned}
\vspace{-4mm}
\end{table*}

\end{document}